\documentclass{article}

\usepackage{PRIMEarxiv}

\usepackage[utf8]{inputenc} 
\usepackage[T1]{fontenc}    
\usepackage{hyperref}       
\usepackage{url}            
\usepackage{booktabs}       
\usepackage{amsfonts}       
\usepackage{nicefrac}       
\usepackage{microtype}      
\usepackage{lipsum}
\usepackage{graphicx}
\graphicspath{{media/}}     

\usepackage{amssymb}
\usepackage{amsmath,amssymb,amsfonts}
\usepackage{amsmath,amsfonts}
\usepackage{algorithmic}
\usepackage{amsthm}
\usepackage{bm}
\usepackage{algorithm}
\def\ci{\noexpand{\perp \!\!\!\!\!\: \perp}}
\newtheorem{dfn}{Definition}
\newtheorem{thm}{Theorem}
  
\title{A Model of Causal Explanation on Neural Networks for Tabular Data
}

\author{
  Takashi Isozaki \\
  Sony Computer Science Laboratories, Inc.\\
  Tokyo, Japan\\
  \texttt{Takashi.Isozaki@sony.com}\\
   \And
  Masahiro Yamamoto \\
  Sony Computer Science Laboratories, Inc.\\
  Tokyo, Japan\\
  \texttt{Masahiro.A.Yamamoto@sony.com}\\
   \And
  Atsushi Noda \\
  Sony Corporation of America \\
  CA, USA\\
  \texttt{Atsushi.A.Noda@sony.com} \\
}

\begin{document}
\maketitle

\begin{abstract}
The problem of explaining the results produced by machine learning methods continues to attract attention. Neural network (NN) models, along with gradient boosting machines, are expected to be utilized even in tabular data with high prediction accuracy. This study addresses the related issues of pseudo-correlation, causality, and combinatorial reasons for tabular data in NN predictors. We propose a causal explanation method, CENNET, and a new explanation power index using entropy for the method. CENNET provides causal explanations for predictions by NNs and uses structural causal models (SCMs) effectively combined with the NNs although SCMs are usually not used as predictive models on their own in terms of predictive accuracy. We show that CENNET provides such explanations through comparative experiments with existing methods on both synthetic and quasi-real data in classification tasks.
\end{abstract}


\section{Introduction}
Research on interpretable machine learning (ML) (e.g., \cite{adadi2018peeking,IF_XAIconcepts}) still attracts much attention. Complex methods, including gradient boosting machines (GBMs) \cite{GBM} and neural networks (NNs), have performed better on many tasks than classical methods such as regression analysis and decision trees. However, methods generally face a tradeoff between interpretability and predictive performance.

Many methods have been proposed to provide explanatory properties to predictive models post-hoc, including research on neural net-specific methods such as Grad-cam \cite{selvaraju2017grad}, and local, model-independent methods such as LIME \cite{LIME}, Anchors \cite{ribeiro2018anchors}, and SHAP \cite{SHAP}. These are being used in industry, in part because they can be easily integrated into the pipeline of the forecasting process using existing ML methods. Nevertheless, several challenges remain in this area, some of which are discussed below.

As a challenge for ML methods that provide explanatory properties, we note that pseudo-correlation and causality are not well addressed. Information about causality is crucial to human understanding of things and situations. The journey of understanding various phenomena in natural sciences such as epidemiology and in social sciences such as psychology is evidence that humans try to understand things through causal relationships (e.g., \cite{MedicineCausal}, \cite{PsyshologyCausal}). Even in our immediate surroundings, cases in which we wish to understand things through causal relationships, such as the causes of crimes and abnormal weather, are an everyday occurrence. If an ML explanatory system that does not consider causality uses a contributing variable to explain its output, the user may interpret the variable as being causally related without being aware of whether it is a pseudo-correlated variable. As a result, incorrect causal knowledge may accumulate in the general public and in local organizations such as marketing and manufacturing companies. In other words, ML and AI risk instilling a false understanding in people. There is also a concern that if the causal relationship is Markovianly not direct, multiple similar explanations will be presented, which will be redundant and difficult to understand. Therefore, it will be effective to present reasons using variables that directly affect the prediction results. This will also increase the likelihood of smoothly leading to intervention actions to change the predicted outcome. In other words, incorporating such structural causality into explainability will not only improve understanding of the reasons for the prediction but also increase the feasibility of a system that seamlessly links prediction and intervention. However, the methods for explaining AI by inferring structural causality among input variables behind the data are still underdeveloped to the best of our knowledge. 

Another challenge is the following: while many explanatory methods, such as LIME and SHAP, assume additivity of multiple variables, there is concern that the predictive model itself is learning more complex patterns among explanatory variables but may be losing some amount of information \cite{NatureMI_Rudin} due to the assumption \cite{kumar2020problems} when providing explanatory properties. One could say that ML should account for complex reasons that frequently occur in reality, but this does not seem to have been well addressed. For instance, the causes of some diseases that are difficult to detect are often thought to arise only when complex factors are combined, and the contribution of a single variable itself is often not high. For example, gender is rarely the sole reason for the occurrence of certain disease A. However, for a given gender, if the people suffer from underlying disease B and have high parameter values for their physical condition, the probability that disease A will co-occur may be very high even though disease B and the high values have only small correlations with disease A. In such a case, these three variables will need to be shown to be the combined reason if disease A is predicted to occur by machines that learned such complex patterns.

This study will present a concept and method and examine their effectiveness in facilitating a causal understanding of predicting models globally and individual prediction results, as well as in dealing with complicated reasoning problems. Specifically, it is a new method for explaining the results of predictions by NNs in tabular data such that causality is assumed to be behind them. Our contributions to this research are as follows. 
\begin{itemize}
\item This method eliminates pseudo-correlations and indirect influencing factors that are difficult to eliminate with many existing methods. As a result, variables that are direct causes can be used as reasons that are easy to understand, and the harmful effects of accumulating erroneous causal knowledge can be reduced. 
\item The method achieves a certain lower entropy bound, indicating that the idea of the proposed method has one important feature in terms of information theory. 
\item In classification-type prediction tasks, the proposed method provides explanations even for patterns where non-additive multiple factors are required for judging individual predictive outcomes. 
\item The proposed method enables single and multiple reasons to be compared using the same new indicator and to evaluate which better explains the classification-based prediction tasks.
\end{itemize}

\section{Related Work}
\label{relatedwork}
According to Arrieta et al.~\cite{IF_XAIconcepts}, the methods to explain AI are mainly divided into transparent models, such as multiple regression, decision trees, and rule-based methods, which are classical and easy to interpret, and post-hoc explainability, being further classified as model-agnostic and model-specific. LIME \cite{LIME} and SHAP \cite{SHAP} fall into the agnostic type: a method that quantifies the impact of specific training data on forecasting results by considering an influence function~\cite{ICML2017best}, in which they linearly approximate relationships among the variables by sampling the neighborhood of the variable (LIME) and utilize the Shapley value of the game theory with the additivity constraint (SHAP). In model-specific terms, numerous methods have been proposed for (deep) NNs (e.g., \cite{TabNetAAAI21}), including decision trees \cite{DecitionTree17DS,FrosstHintonDT17}. A derivative of an NN by Agarwal et al. \cite{NAM} is a linear model but characterized by its ease of application to real data. Tsang et al. \cite{TsangDetectInteraction} proposed a method to understand the non-additive coupling of neurons in NNs, and there are many other efforts to understand the behavior of the middle layers of NNs (e.g., \cite{Layer-Wise_NN}), but those are not from causal perspectives on tabular data. 

Complex but highly accurate methods such as NNs and GBMs are being used in many situations for predicting tasks, including by practitioners. Among the deep learning techniques applicable to tabular data, many derivatives have been proposed, such as ResNet \cite{ResNet}, SNN \cite{SNN}, TabNet \cite{TabNetAAAI21}, and FT-Transformer \cite{NeurIPS21DNNtabular}. On the other hand, several methods including XGBoost \cite{XGBoost} and LightGBM \cite{LightGBM} have been recognized as highly accurate GBMs. Comparison experiments have been conducted between NNs and GBMs on prediction performance for tabular data \cite{NeurIPS21DNNtabular,NeurIPS22WhyTreeBasedOutperform,DNN_survey_IEEE2022}. In many cases, NNs and GBMs differ little in terms of accuracy \cite{WhenNNoutperformGBDT23,tabular_NN24}, and the superiority or inferiority seems to depend on the dataset and NN-structure and hyperparameter setting \cite{NeurIPS22WhyTreeBasedOutperform,KadraNeurIPS2021welltuned}. Debate continues about the conditions under which NN or GBM will produce better results \cite{WhenNNoutperformGBDT23}.

The viewpoint of causality has begun to be actively discussed in NN models and explanations of ML. Chattopadhyay et al. \cite{NNasSCM19} presented the idea of interpreting the NN model itself as a structural causal model. Problems related to causality even in highly evaluated methods such as SHAP have been addressed \cite{kumar2020problems}. 
A case study also indicates that problems may arise in explaining SHAP's predicted results without taking causality into account \cite{InterpretPredCausal-FLAIRS23}. 
Incorporating causal knowledge and partial causal order has been investigated \cite{FryeFeigeRowat20nips}, and Shapley values that incorporate and consider causality have been proposed \cite{CausalShapley20nips}. Janzing et al. \cite{JanzingMinoricsBlobaumAISTATS20} also discussed Shapley values from a causal point of view. However, the non-additivity of multiple explanatory variables is difficult to handle because Shapley values are used. Schwab and Karlen \cite{CXPlain19} proposed a method to estimate the feature importance using Granger's causality and has a feature of uncertainty estimation in particular. However, this approach makes it difficult to identify whether a correlation is a Markovianly direct causation, indirect causation, or pseudo-correlation among input variables. Narendra et al. \cite{IBM_DLCM} and Harradon et al. \cite{DNN_Causal} used causal models as relational models of the entire NNs including intermediate layer neurons, and their goals are thus different from ours. 
Ahmad et al.~\cite{WACV_causal_2024} studied how to probe and interpret important hidden layer neurons by intervening in the weights of NNs. 

\section{Methodology}\label{methodology}
\subsection{Preliminaries}
Correlation does not necessarily mean causation as is often said, and to extract causally explaining variables from input variables for prediction, we use structural causal models (SCMs) \cite{PearlCausality,PC}. For instance, a pseudo-correlation (one of the correlations) is a relationship between two resultant variables that arises from another causal variable, which we wish to distinguish from causation. An SCM is a graphical representation of causal relationships with nodes and edges, where nodes represent random variables and edges represent causal relationships. The model is usually assumed to be a directed acyclic graph (DAG), which is also assumed in this study. The model discovery methodology is based on the axiom, named causal Markov condition (or assumption)~\cite{PearlCausality,PC,SpirtesIntro} (see Appendix~\ref{SCMbase}), which is described with conditional independence, and only the oriented edges of the models denote Markovianly direct causations. Thus, the absence of an edge between two variables in the SCM inferred from observational data means that there is no correlation or zero partial correlation in combination with a given set of variables from the viewpoint of statistics~\cite{PC}. Conversely, the presence of an edge means that there is no set of variables in the data that would have zero partial correlation. Therefore, finding an SCM in a dataset means performing a partial correlation analysis and decomposing it into correlations including uncorrelated, pseudo-correlated, and indirect causality (which we call indirect correlation), and direct causality or pseudo-correlations generated by hidden variables or selection bias~\cite{PC} (which we call direct causality or direct correlation).

Here, we focus on an important mathematical property of structural causality inferred on the basis of partial correlations, which consists of a relationship between two variables that are partially correlated. This property is involved and often referred to in the context of feature selection with explanatory properties for predictions in this study. We thus refer to the special bivariate relationship of non-zero partial correlation in a given dataset, which is related to structural causality in the context of this study as follows (see Appendix~\ref{SCMbase} for references to terminologies).
\begin{dfn}
In a dataset $\mathcal{D}$, let $X$ and $Y$ be random variables, and $\bm{Z}$ be a set of random variables. If there is no variable set $\bm{Z}$ that makes the partial correlation zero for $X$ and $Y$ in $\mathcal{D}$, and $X$ precedes $Y$ in causal order, then $X$ is said to have a characteristic correlation with $Y$ in $\mathcal{D}$, and $X$ is also called the characteristic correlated variable of $Y$, denoted by $\text{CCV}(Y)$.
\end{dfn}
We note the Shannon entropy and its conditional one~\cite{CoverThomas} as the preliminary. Let $X$ be a discrete random variable and $\bm{Y}$ be a set of discrete random variables. The Shannon entropy $H(X)$ and conditional entropy $H(X\,|\,\bm{Y})$ are defined as~\cite{CoverThomas}
\begin{align}
&H(X) = - \mathbb{E}_{P(X)} \log P(X),\\
&H(X\,|\,\bm{Y}) = -\mathbb{E}_{P(X,\bm{Y})} \log P(X\,|\,\bm{Y}).
\end{align}
The characteristic correlated variables (CCVs) have an important meaning from the viewpoint of information theory, as summarized by the following property with the entropy. Its proof is given in Appendix~\ref{ProofTheorem}. 
\begin{thm}\label{Th_entropy}
In a given dataset $\mathcal{D}$ with a discrete random variable set $\bm{V}$, if there is a causal DAG for the set $\bm{V}$, the entropy of $X \in \bm{V}$ subject to the variable set CCVs of $X$ (CCV ($X$)), $H(X\,|\,\text{CCV}(X))$, achieves the lower limit of the conditional entropy given any set $\bm{S}$ such that $\text{CCV}(X) \subseteq \bm{S} \subseteq \bm{V}\backslash X$ and $X$ is not an ancestor of $\bm{S}$. That is,
\begin{equation}
\text{min} \, H(X\,|\,\bm{S}) = H(X\,|\,\text{CCV}(X)).
\label{CCV_entropy}
\end{equation}
\end{thm}
CCVs have unique explanatory abilities because they can possess non-zero partial correlations with a target variable in a dataset. We will henceforth often use the terms characteristic correlation and CCV instead of the mathematically vague terms causality and causal variables because different approaches to causality and inference exist.

\subsection{The Architecture}
\label{architecture}
\begin{figure}[tb]
\centering
\vspace*{1.5cm}
\hspace*{-2.0cm}
\includegraphics[bb = 0.000000 0.000000 1113.758121 618.379509, width = 7.0cm]{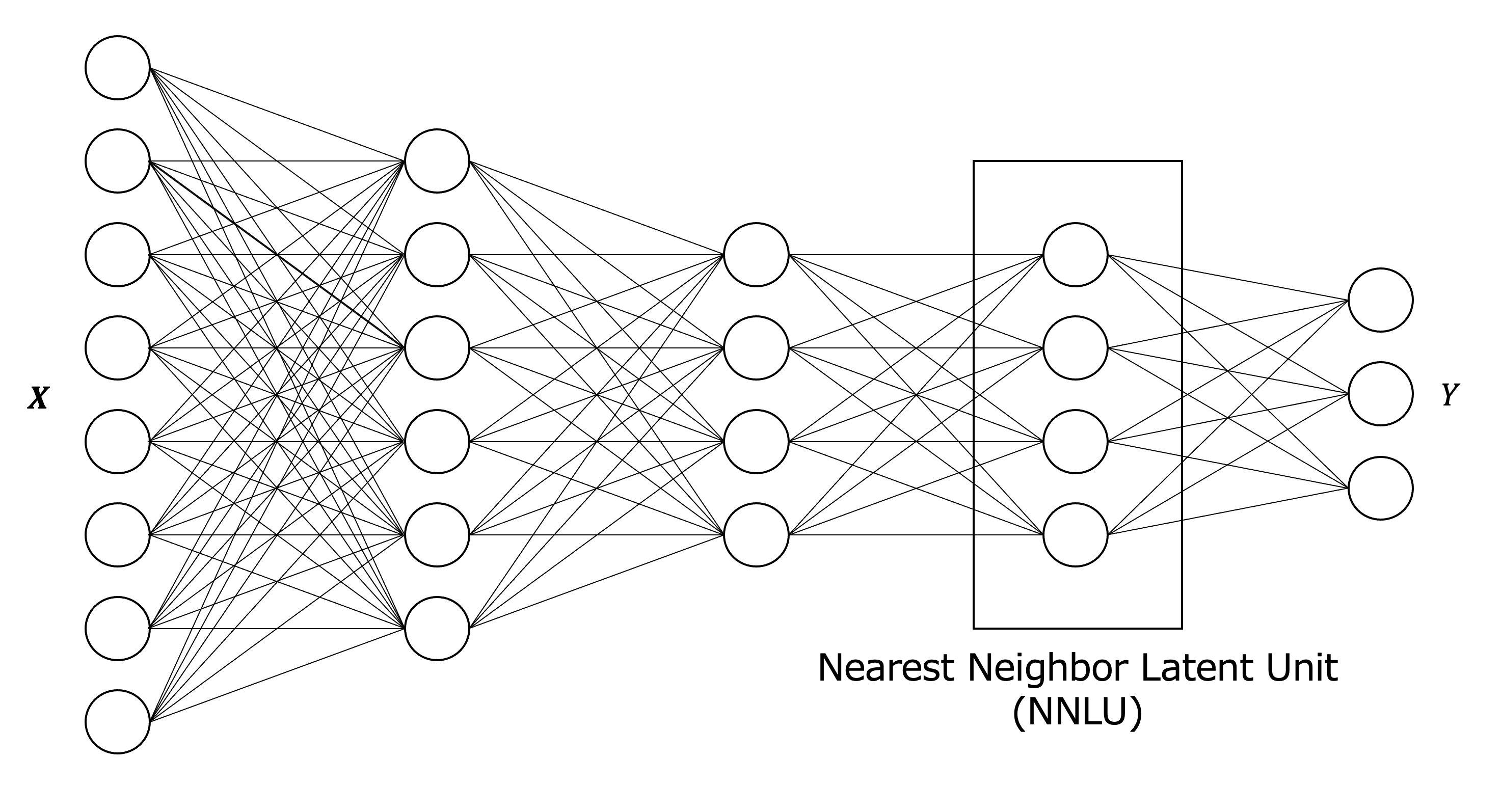}
\vspace*{-0.1cm}
\caption{The nearest neighbor latent unit (NNLU) to output variable $Y$ in a neural network with input variable set $\bm{X}$. The proposed method causally analyzes the neurons in the NNLU. }
\label{NNLU-network}
\end{figure}

We present Causal Explanations for Neural NETwork predictors (CENNET). This research seeks to develop an integrated architecture that will allow NNs to understand how they learned (i.e., global explanation) and develop mechanisms to exploit the global properties to explain predicted outcomes (i.e., local explanation). Since the final output layer of an NN is determined in many models by the weighted average of the outputs of the neurons in the layer before it, we consider that analyzing not only the final output layer but also the layer before it will contribute to a deeper understanding of learned NNs. Therefore, in this study, we name the layer adjacent to the final output variable as the nearest neighbor latent unit (NNLU) and analyze the training results of this layer. Figure~\ref{NNLU-network} illustrates the NNLU in a NN. On the basis of the assumption that a causal relationship exists between the explanatory variables in the tabular data and the final output variable, we consider the outputs of the neurons in the NNLU as random variables and analyze the partial correlation and causality of each neuron. Since the neurons in the NNLU are constructed by learning from the input variables, the input variables are obviously causes for the NNLU neurons. Specifically, one causal model is inferred among the input variable set and one neuron in the NNLU, and other causal models are inferred for each other neuron in the NNLU. The causal analyses thus allow us to figure out which variables are CCVs to each neuron in the NNLU.

At the same time, this architecture is expected to avoid the drawback of the graphical models that using the ordinary graphical models as predictive models, except for naive Bayes and its extensions, is not a good choice in terms of predictive accuracy, as found in the work of Friedman et al.~\cite{FGG}. That is, while the number of contributing variables is not sufficient when extracting causal variables that directly impact the final output layer, CENNET is expected to extract a larger number of contributing and directly causal variables because it looks for the CCVs for each neuron in the NNLU. The effectiveness of the architecture proposed in this study, which focuses on the causal analysis of NNLUs, will be discussed later with our preliminary experiments. 

\subsection{Global Explanation with Causal Analysis}
Based on the architecture of CENNET proposed in the previous subsection, 
the following is the concrete method for globally understanding how each neuron in the NNLU is trained, by performing a CCV selection-analysis of the neurons in the NNs. Let the output variables of NN be $\bm{Y}$, the set of input explanatory variables be $\bm {X}$, and their instances be $\bm {y}$ and $\bm{x} $. Let $\bm{n} \, (= \{n_ {1},\dots,n_{p} \})$ be the neuron group in the NNLU to $\bm{Y}$ ($p$ is the number of neurons in the NNLU). Here, note that $\bm{X}$ precede $n_{i}$ in the causal order. We assume that there is an SCM for the relationships between each NNLU neuron and explanatory input variables. These NNLU neurons are analyzed by using the following procedure. For each neuron $n_ {i}$ in the NNLU, a set $\bm{X}_{E}(n_ {i})$ that denotes the CCVs in the input variable $\bm{X}$ is extracted with a causal search algorithm. $\bm{X}_{E}(n_ {i})$ is regarded as a set of important features of the NNs.

\begin{figure}[t]
\centering
\vspace*{-17.0cm}
\hspace*{1.6cm}
\includegraphics[bb = 0.000000 0.000000 4872.000 1870.000, width = 60.0cm]{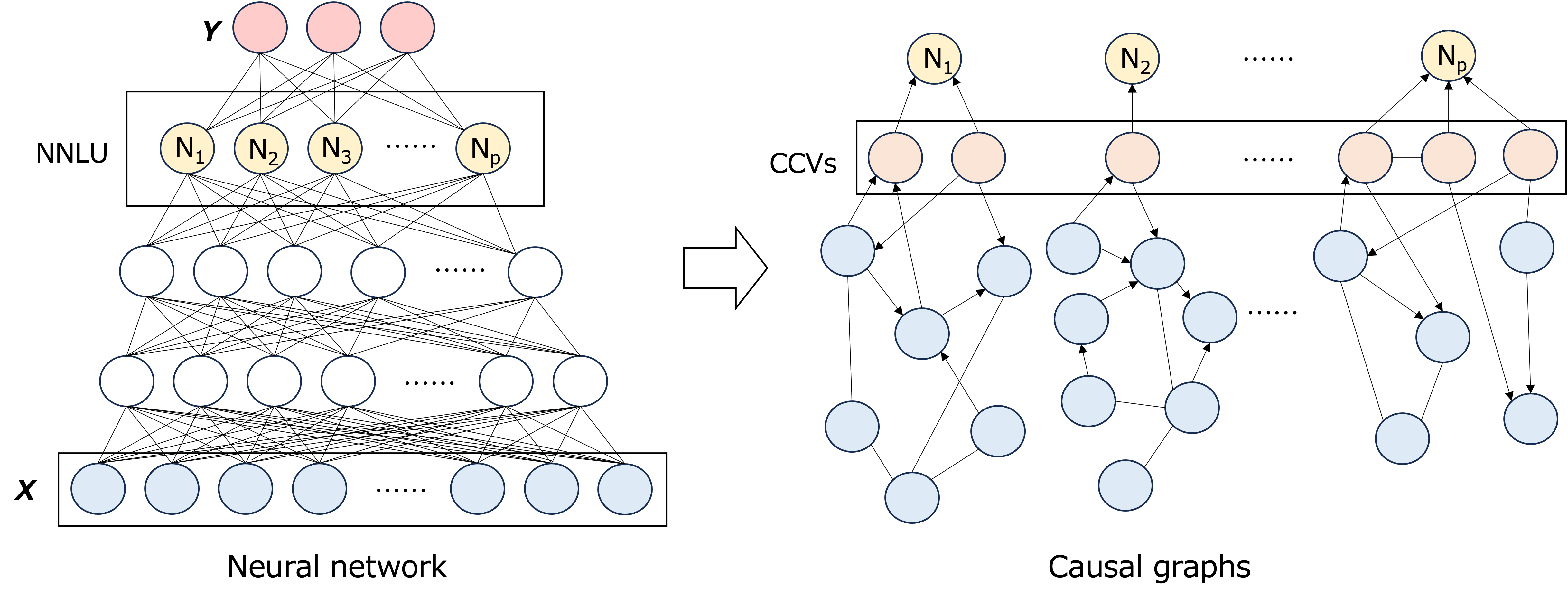}
\caption{The overview of CENNET method. In an NN model, CENNET extracts $p$ NNLU neurons and infers $p$ causal models consisting of input variables $\bm{X}$ for each NNLU neuron.}
\label{overview}
\end{figure}

If the input variables are continuous variables, regarding the relationship between $n_{i}$ and $\bm{X}_{E} = \{X_{Em} \} (m = 1, \dots, k)$ ($k$ represents the number of CCVs in $ n_{i}$), for example, an explanatory model of $\hat{n}_{i }$ of the NNLU is expressed by the causal linear structural equation~\cite{PearlCausality}, with an assumption of linear models, as $\hat{n}_{i}=\sum_{m=1}^{k} \alpha_{m} X_{Em} + \epsilon_{m},$ where $\alpha_{m} $ denotes the coefficient and $\epsilon_{m}$ denotes a noise term. In contrast, if the input variables are categorical, an explanatory model is generated by conditional probability distributions. For the former case, the functional type needs to be defined by the aforementioned linear structural equation model, so various nonlinearities and the combined effect of two or more variables are not easy to express, especially for neurons generated and trained. Therefore, in the following, both the input variables and the output of the trained $n_{i}$ neuron will be treated as discretized variables in the causal analyses. This model can then be examined in detail by $P(n_{i}\,|\,\text{CCV} (n_{i}))$ with CCV~\cite{PearlCausality}. It is important to add that $H(n_{i}\,|\,\text{CCV}(n_{i}))$ achieves the lower bound subject to only CCVs by Theorem~\ref{Th_entropy}. 
CENNET thus enables us to understand how the NNLU neurons are trained on the basis of input variables. The final output $\bm{Y}$ can then be globally explained using CCVs and the weights between $Y$ and the neurons. 
The computational complexity is the number of neurons in the NNLU multiplied by the computational complexity for estimating the causal graphs, but since the part concerning the number of neurons can be completely parallelized, it is actually the computational complexity for estimating the causal graphs. 
Figure~\ref{overview} shows the overview of the CENNET method for global understanding of NNs. For NNs, CCVs are extracted by inference of SCMs.

\subsection{Local Explanation with NNLU Neurons and CCVs}
We also propose a methodology for providing reasons for individual predictions as a function of CENNET. Hereafter, we focus on the prediction of classifications: specifically, we provide a theory for the binary output type and then extend it to a multi-class one. Let $\bm{x}$ and $\hat{y}$ be instances of input attributes and the final output of predictive variable $Y$ in a trained NN, respectively. $\hat{y}$ is then expressed by
\begin{equation}
\hat{y}= \sum_{i}^{p} w_{i} \hat{n}_{i} + b \label{basicNN},
\end{equation}
where $p$, $\hat{n}_{i} (i=1, \dots , p)$, $w_{i}$, and $b$ denote the number of NNLU neurons, output values of the neurons, the weights between $Y$ and $n_{i}$, and a constant bias, respectively. We obtain the classification categories by inputting the values $\hat{y}$ into some function like the sigmoid one in NNs. The characteristic correlation analysis is first made for trained NNs, and then CCVs are extracted for each NNLU neuron. We can calculate the indices for accountability in each instance, as we explain in the following.

We here formalize a mechanism that provides the accountability index, which is a unification of the single variable and multivariate weights of CCVs, by discretizing all the related variables including $\hat{y}$, $\hat{n}_{i}$, and input $\bm{x}$. Let the weight of CCVs be an element of entropy-based explanation powers (EEPs), which is defined with marginal and conditional entropy elements as
\begin{equation}
\text{EEP}(z,\bm{x}_{e})=\log \{P(z\,|\,\{\bm{x}_{e} \}) / P(z)\},
\label{EEP}
\end{equation}
where $z$ denotes a value of the target variables and $\bm{x}_{e}$ denotes a configuration of CCVs $\bm{X}_{E}$. The number can be restricted when computing Eq. (\ref{EEP}) for combinations of CCVs by calculating the expectation value of an EEP (extended mutual information (EMI)) for the maximum size of $\bm{X}_{e}$ determined in advance. The EMI is defined as follows.
$\text{EMI}= H(X) - H(X\,|\,\bm{X}_{e}). $ 
Next, we define the index of total power for explanation using an EEP. In trained NNs, it is known whether the NNLU nodes have positive/negative contributions due to the codes of their weights $w_{i}$. Therefore, the positive/negative set of the NNLU nodes can be defined as $n_{+} \subseteq n: w_{i} \geq 0$, $n_{-} \subseteq n: w_{i}<0$. Here, we assume that any neuron in the NNLU has a value of zero or more, which we determine by using a function such as ReLU activation. Let $\tilde{n}_{i}$ be a discretized value of $\hat{n}_{i}$ and EEP ($\tilde{n}_{i}, \bm{x}_{e})$ be its EEP value. We propose an explanatory power by combining the neurons' weight contribution to predictions with the information-theoretic element: EEP. We formulate two quantities, which we call \textit{positive explanation power} (PEP) and \textit{negative explanation power} (NEP), and then define a unifying index of explanations, which we call \textit{total explanation power} (TEP), as follows.
\begin{align}
\text{PEP}(\bm{x}_{e}) &= \sum_{n_{i} \in n_{+},\text{EEP}>0} w_{i} \hat{n}_{i} \cdot \text{EEP}(\tilde{n}_{i},\bm{x}_{e}) + b \label{PEP}\\
\text{NEP}(\bm{x}_{e}) &= \sum_{n_{i} \in n_{-},\text{EEP}>0} w_{i} \hat{n}_{i} \cdot \text{EEP}(\tilde{n}_{i},\bm{x}_{e}) + b \label{NEP}\\
\text{TEP}(\bm{x}_{e}) &= \text{PEP}+\text{NEP}. \label{TEP}
\end{align}

In this research, we focus on NNLU neurons that have often learned non-linear complex patterns from multivariate input data because the entanglement is probably a source of the excellent performances of NNs in various predictions. 
CENNET can cite explanatory reasons including this multi-variable coupling effect, which are evaluated by using an index of explanatory power that combines the values of NNLU neurons with weights and entropy-like quantities. 
In addition, we use CCV-based analysis to remove pseudo-correlative and similar variables when selecting reasons and also to give causal reasons. These are the unique characteristics of CENNET. Additionally, note that CENNET can be applied to various types of NNs, although we use only the multi-layer perceptrons (MLPs) in our experiments described in the next section. 
We show the all-over CENNET algorithm for the global and local explanations in Algorithm~\ref{alg:algorithm}.
If the number of neurons in the NNLU is $l$, the number of combinations consisting of CCVs is $p$, and the number of samples to be explained is $t$, the computational complexity here is of $\mathcal{O}(lpt)$ since the EEP calculation can be precomputed and cached from the training data.

\begin{algorithm}[tb]
 \caption{CENNET algorithm for both global and local explanations}
 \label{alg:algorithm}
 \textbf{Input}: A trained NN model, input variables $\bm{X}$, prediction variable $Y$, a training dataset $\{\bm{D}\}$ for $\{ \bm{X}, Y\}$, and an instance set $\{\bm{d}\}$ for $\bm{X}$\\
 \textbf{Parameter}: $m$: the maximum order of multivariate explanations for output\\
 \textbf{Output (Global)}: List of CCVs for the NN predictor\\
 \textbf{Output (Local)}: List of explanatory instances for the prediction value in $Y$ in $\{\bm{d}\}$\\
 \textbf{Global Explanation}:
 \begin{algorithmic}[1] 
 \STATE Extract a set of neurons in NNLU from the NN. 
 \FORALL {neuron $n_{i}$ in NNLU $( 1<= i <= p)$}
 \STATE Infer a causal model among the set $\{Y, \bm{X}, n_{i}\}$ from $\{\bm{D}\}$.
 \STATE Extract CCV($n_{i}$).
 \ENDFOR
 \STATE \textbf{return} $\{ \text{CCV}(n_{1}), \dots, \text{CCV}(n_{p}) \}$
 \end{algorithmic}
 \textbf{Local Explanation}:
 \begin{algorithmic}[1] 
 \STATE Let $\bm{X}_{n}\,\leftarrow\,\, \bigcup_{\text{all}\,\,n_{i}} \text{CCV}(n_{i})$
 \STATE Extract $\bm{x}_{n}$: the values of $\bm{X}_{n}$ in $\{\bm{d}\}$.
 \FORALL {neuron $n_{i}$ in NNLU}
 \STATE Extract $w_{i}$: the value of a weight between $Y$ and $n_{i}$.
 \STATE Calculate $\hat{n}_{i}$: the value of each $n_{i}$.
 \ENDFOR
 \FORALL {configurations $\bm{x}_{e}$ in $\bm{x}_{n}$ until its size $<= m$}
 \STATE Calculate TEP($\bm{x}_{e}$) by using Equation (\ref{TEP}).
 \ENDFOR
 \STATE Sort $\bm{x}_{e}$ by values of TEP.
 \STATE \textbf{return} A list of sorted $\bm{x}_{e}$
 \end{algorithmic}
\end{algorithm}

Finally, two extensions are worth mentioning. 
We discuss the case of multi-class prediction. Suppose that $\bm{y}$ is the final output set of variables and $\bm{y}\in \mathbb{R}^{d}$. Let $m$, $\bm{N}$, $\bm{W}$, and $\bm{b}$ be the number of nodes in an NNLU, their $m$-dimensional vector, the weight $d \times m$ matrix, and the $d$-dimensional bias vector, respectively. Equation (\ref{basicNN}) then changes into its multivariate version $\bm{y}=f(\bm{W}\bm{N}+\bm{b})$ for multi-class prediction. This is the straightforward extension, where $f$ is similar to the softmax function.
Next, we mention the computability for NN models that are not small. 
CENNET analyses each NNLU neuron for an NN using causal discovery algorithms and the analyses can be parallelized. In addition, tabular datasets with causal structures behind them usually do not require very large NNs. That is, the number of NNLU neurons is not large. Thus, the issue of computational cost is not as serious.

\section{Experiments}
\label{Synthetic}
In this section, we present the results of our quantitative experiments. In this experiment, three synthetic datasets (Non-Linear Additive, Non-Linear Non-Additive, and Category experiments) were prepared with reference to the method of Chen et al.~\cite{pmlr-v80-chen18j} and four quasi-real-world datasets were used to compare with four existing methods: LIME \cite{LIME}, Stabilized-LIME (S-LIME) \cite{SLIME}, SHAP \cite{SHAP}, and Active Coalition of Variables (ACV) \cite{ACV}. 

The experimental process is described below. 
We defined important variables as those that directly affect the prediction variable and were known in advance, 
trained the MLPs with two hidden layers, used the MLPs for predictions and 
applied each XAI method to the predictions. 
Then we calculated the importance of the explanatory variables for each sample and ranked the variables on the basis of their importance. Finally, we evaluated each method by using its average rank of the important variables for all test samples. The smaller this rank is, the more highly the method is evaluated. 
The details of the datasets are as follows, and also in Appendix~\ref{ExpApp} for supplementary setting information. 
\\
{\bf Non-Linear Additive experiment:} We generated values of a set of 10 explanatory variables $\bm{X}=\{X_{1},\dots, X_{10}\}$ from a 10-dimensional standard Gaussian distribution. The value of the binary prediction variable $Y$ was generated by following a non-linear and additive form $P(Y=1\,|\,\bm{X}) \propto \exp(\sin(0.2 X_{1}) + 0.1 |X_{2}|+ X_{3} + \exp(-X_{4}))$. In this dataset, $X_{1}, X_{2}, X_{3}$, and $X_{4}$ were defined as the important variables and the others were not, and thus the methods that output high importance for these four variables were evaluated as good ones. 
We calculated the average ranking of the important variables in all samples for each method and compared the methods on the basis of their average ranking.
\\
{\bf Non-Linear Non-Additive experiment:} In the set of variables $\bm{X}=\{X_{1},\dots, X_{10}\}$, the values of input variable $X_{1}$ were generated from the uniform distribution in the range from -10 to 10 and other values of each variable in $\bm{X}\backslash X_{1}$ were generated from the standard Gaussian distribution. The values of prediction variable $Y$ were generated by non-linear and non-additive functions as follows, where $P := P(Y=1\,|\,\bm{X})$. 
If $X_{1} > 7.0$, $P \propto \sin(X_{2})$, else if $X_{1} > 4.0$, $P \propto \cos(X_{3})$, else if $X_{1} > 0.0$, $P \propto \tan(X_{4})$, else if $X_{1} > -4.0$, $P \propto \exp(2X_{5})$, else if $X_{1} > -7.0$, $P \propto \tanh(X_{6})$, else $P \propto $ sin$(X_{7})$.
The pair of $X_{1}$ and another variable was an important one depending on the values of $X_{1}$ in this dataset (e.g., if $X_{1}$ is larger than 7.0, the important pair is $X_{1}$ and $X_{2}$). Thus, a method that outputs high importance for the known important pairs was evaluated as a good one. 
Specifically, first, for each sample in the test data, the importance of each pair of explanatory variables was calculated, where there are a total of $C(10,2)=45$ possible pairs. The variable pairs were then ranked in order of their importance, and the ranking of important variable pairs was calculated. 
\\
{\bf Category experiment:} We generated the values of explanatory variables $\bm{X}$ as 10-dimensional categorical ones. The value of the prediction variable $Y$ was generated from the following categorical and non-additive distributions, where $P := P(Y=1\,|\,\bm{X})$, and the sets are denoted by the form $\{X_{1}, X_{2}, X_{3}, P\}$. 
Those are defined as $\{0, 0, 0, 0.80\}, \{0, 1, 0, 0.70\}, \{0, 1, 1, 0.20\}, \{0, 0, 1, 0.25\}, \{1, 0, 0, 0.20\}, \{1, 1, 0, 0.80\}, \{1, 0, 1, 0.85\}, \{1, 1, 1, 0.20\}$. 
In these experiments, $X_{1}$, $X_{2}$, and $X_{3}$ were defined as the important variables. 
The methods that were able to assign higher levels of importance to the combination of these three variables are evaluated more highly.
\\
{\bf Quasi-real data experiment:} 
We would like to evaluate the proposed method on real-world data, but there are few real data with many variables for which causal relationships are known. We thus evaluate the method using the following quasi-real-world data. 
We used datasets generated by sampling from four expert knowledge models (Alarm, Hailfinder, Insurance, and Carpo (see Appendix~\ref{QRWD} for the details)) with specified probability distributions and directed graphs and used 10 variables with binary categories as prediction variables. 
Since a directed graph was set up for each dataset, the variables that could be considered as direct causes of each prediction variable were known in advance. In this experiment, these directly causal variables were defined as important variables, and as in the case of the synthetic data experiment, each method was evaluated on the basis of the average ranking of the importance of that set of variables. We can test whether each method can remove pseudo-correlated and indirect causal variables from the explanation of predictions with this experiment.

The experimental procedures are as follows. 
For experimental and implementation details, please see also Appendix~\ref{ExpApp}. 
The number of datasets for each dataset was 10,000, split into training, validation, and test data.
The validation and test sets were used for hyperparameter tuning and evaluation, respectively. 
The number of neurons was set to 5 in the NNLU and 16 in another hidden layer. 
After the MLPs were trained, the data of each numerical variable were split into three categories with equal frequencies and used as categorical data for CENNET. 
We implemented and used a modified version of PC algorithm~\cite{PC} for inference of causal models. 
Then, we calculated the importance of explanatory variables using Equations (\ref{PEP}), (\ref{NEP}), and (\ref{TEP}) for CENNET. 
We used the official Python implementations for LIME, S-LIME, SHAP, and ACV to produce the experimental results. 
In the Non-Linear Non-Additive, Category, and Quasi-real data experiments, we calculated the combination importance on the basis of the sum of variables for the existing methods.

\begin{figure*}[t]
\centering
\vspace*{-33.0cm}
\includegraphics[bb = 0.00 0.00 3057.00 1695.00, width = 78.0cm]{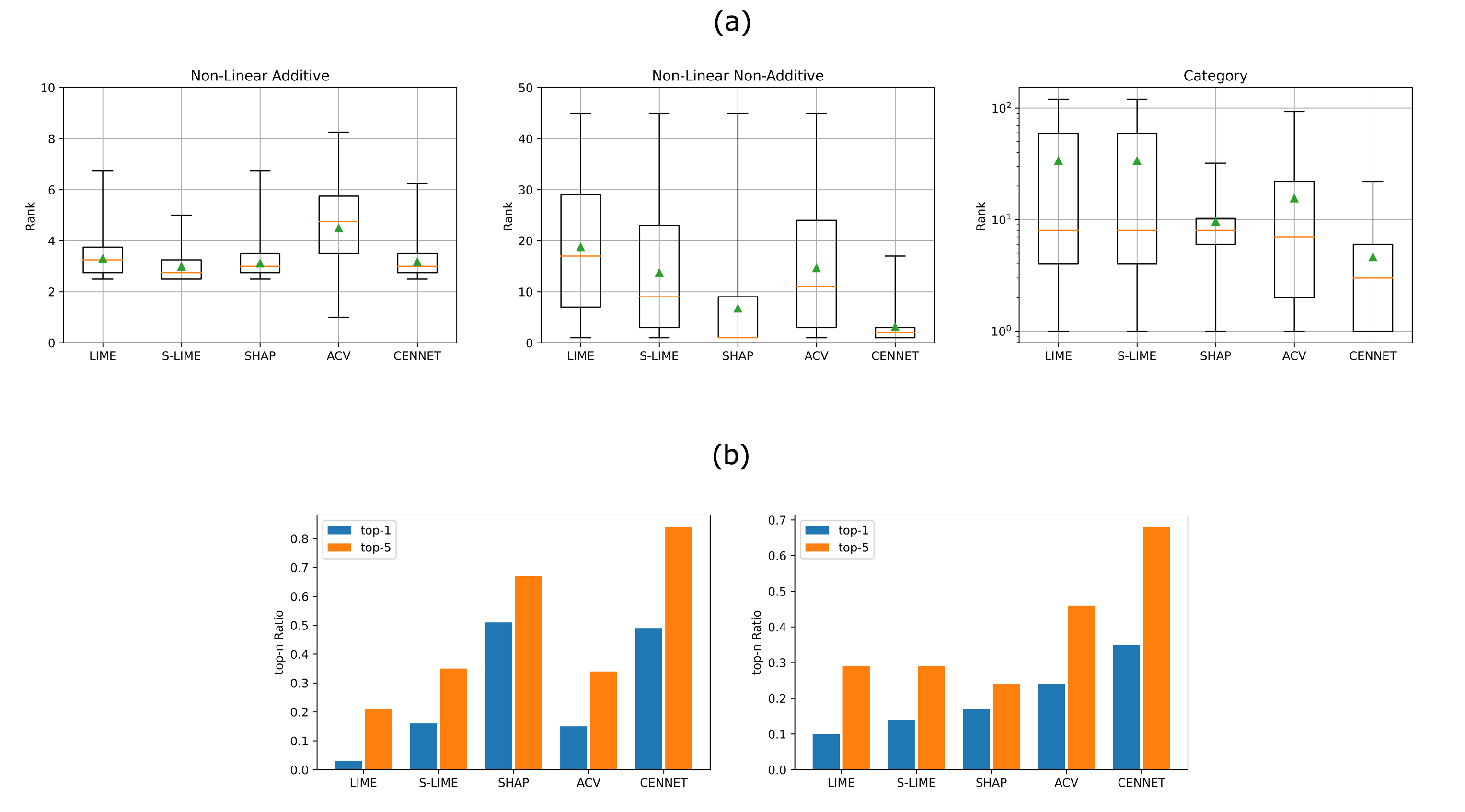}
\caption{(a) Box plots for three experiments. Left: Non-Linear Additive. Middle: Non-Linear Non-Additive. Right: Category. Orange lines and green triangles in each box denote the median and mean, respectively. Lower average ranks are better. (b) Bar plots for two experiments. Left: Non-Linear Non-Additive. Right: Category. Blue and orange bars denote the top-1 and top-5 ratios, respectively.}
\label{box_bar_plot_res}
\end{figure*}

Figure~\ref{box_bar_plot_res} (a) and (b) show the results of the experiments with synthetic data. 
The left panel of Figure~\ref{box_bar_plot_res} (a) shows a box plot of the average ranking of the important variables (IVs) $\{X_{1}, X_{2}, X_{3}, X_{4}\}$ in the Non-Linear Additive experiment. 
The middle and right panels of Figure~\ref{box_bar_plot_res} (a) show the same for the IV sets in both the Non-Linear Non-Additive and the Category experiments, respectively. 
The left side of Figure~\ref{box_bar_plot_res} (b) shows the percentage of IV pairs that could be ranked as the top-1 and within the top-5 in the Non-Linear Non-Additive experiment, and the right side shows the same for the IV sets in the Category experiment.

The left panel of Figure~\ref{box_bar_plot_res} (a) shows that in the Non-Linear Additive experiment, CENNET performed almost the same as the existing three methods (LIME, S-LIME, and SHAP) and that ACV was the worst. 
The middle and right panels of Figure~\ref{box_bar_plot_res} (a) show that CENNET outperforms all the existing methods both in the Non-Linear Non-Additive and Category experiments, despite using a simple method for discretization in CENNET. It can be said that CENNET statistically provides better explanations of the predictions for data that have combinatorial explanatory variables influencing the prediction variable and therefore require a high level of explanatory power. The left panel of Figure~\ref{box_bar_plot_res} (b) shows that in the top-1 ranking of important variable pairs in the Non-Linear Non-Additive experiment, CENNET is at the same level as SHAP statistically since no significant difference was seen in the McNemar test, where $p=0.324$. The percentage of important variable pairs that are within the top-5 ranking is higher for CENNET than for the other existing methods. In addition, the right panel of Figure~\ref{box_bar_plot_res} (b) shows that CENNET outperforms all others in both the top-1 and top-5 ranking in Category. These results indicate that CENNET statistically ranks important variable sets stably higher than the existing methods.

Table \ref{res_realdata} shows the results of the Quasi-real data experiment. Each value denotes the average rank and its standard deviation. 
The lower values are better, and bold letters represent the best results examined by t-tests with 1$\%$ significance level in each setting. 
CENNET outperformed the existing methods in seven out of ten experimental settings. This shows that CENNET is able to present more direct causal variables of the prediction variable as explanations for the prediction than other methods.

\begin{table*}[htb]
\caption{Results of Quasi-real data experiments. Numerical values denote the average ranking of direct causal variables of the prediction variables in terms of the importance. Lower is better, and bold represents the best results in each setting.}
\begin{center}
\scalebox{0.95}[0.95]{
\small{
\begin{tabular}{lccccc}
\hline
  Dataset (Prediction Variable)&   LIME  & S-LIME & SHAP & ACV & CENNET  \\ \hline
Alarm (SHUNT) & 6.96 $\pm$ 3.50 & \textbf{6.93 $\pm$ 3.39} & 37.50 $\pm$ 10.27 & 22.41 $\pm$ 12.49 & 28.31 $\pm$ 9.19\\ 
Alarm (HISTORY) & \textbf{1.00 $\pm$ 0.00} & \textbf{1.00 $\pm$ 0.00} & 1.64 $\pm$ 0.85 & 3.26 $\pm$ 1.71 & 3.79 $\pm$ 0.84 \\
Alarm (CATECHOL) & 103.18 $\pm$ 69.98 & 102.12 $\pm$ 68.05 & 96.74 $\pm$ 96.11 & 118.11 $\pm$ 87.88 & \textbf{24.75 $\pm$ 30.08} \\
Hailfinder (WindFieldMt) & 7.69 $\pm$ 4.92
 & 7.65 $\pm$ 4.94
 & 7.54 $\pm$ 4.22 & 7.11 $\pm$ 4.10 & \textbf{2.13 $\pm$ 2.63} \\ 
Hailfinder (ScenRelAMCIN) & 9.06 $\pm$ 4.49 & 9.25 $\pm$ 4.52 & 7.72 $\pm$ 3.76 & 7.96 $\pm$ 5.67 & \textbf{1.43 $\pm$ 0.94} \\ 
Insurance (Othercar) & 4.32 $\pm$ 4.64 & 4.30 $\pm$ 4.60 & 3.49 $\pm$ 3.23 & 4.86 $\pm$ 3.60 & \textbf{2.28 $\pm$ 2.26} \\ 
Insurance (Vehicleyear) & 79.21 $\pm$ 19.65 & 78.36 $\pm$ 18.80 & 70.72 $\pm$ 16.17 & 43.74 $\pm$ 29.26 & \textbf{29.14 $\pm$ 5.95} \\ 
Insurance (Antilock) & 41.22 $\pm$ 12.61 & 41.15 $\pm$ 12.52 & 15.89 $\pm$ 13.96 & 31.86 $\pm$ 17.97 & \textbf{8.63 $\pm$ 6.99} \\ 
Insurance (Airbag) & 12.69 $\pm$ 6.27 & 12.80 $\pm$ 6.34 & 4.74 $\pm$ 1.71 & 13.35 $\pm$ 7.17 & \textbf{3.54 $\pm$ 4.08} \\
Carpo (N42) & 111.42 $\pm$ 22.45 & 111.61 $\pm$ 22.28 & 95.36 $\pm$ 30.02 & \textbf{81.69 $\pm$ 42.58} & 86.23 $\pm$ 29.11\\
\hline
\end{tabular}
}
}
\label{res_realdata}
\end{center}
\end{table*}

\section{Discussion}
\label{discussion}
For complex prediction models, many existing methods with explanatory properties quite possibly explain predictions by distorting the learned models, as pointed out by Rubin~\cite{NatureMI_Rudin}. 
The main ways of distorting the learned contents seem to be linear and additive approximations, and projections to lower dimensions, for ease of explanations or human comprehension. Our method is designed to avoid those two approximations, especially for classification tasks, and we showed its effectiveness in our evaluations in the previous section. 
In addition, CENNET probably projects learned patterns to some lower dimensions, but the way seems good as we explain below. 
The auto-feature generation function of NNs is valuable for highly predictive performances because it combines input variables to generate excellent features that are difficult for humans to create. However, humans have difficulty understanding how the prediction is done due to the auto-generation property, and thus these features should be projected onto an information space so that humans can understand. In this study, for this purpose, we decompose the created features in the NNLU into CCVs consisting of input variables, resulting in proper dimensional reductions from the viewpoint of Shannon entropy. The CCV dimensions are often much lower than the ones of input variables due to eliminating spurious correlations and indirect causations, which are not always necessary for explanations. Thus, we consider that CENNET using causal inference with CCVs has preferable properties for helping humans understand predictions of NN models and the models themselves globally.

It is worthwhile mentioning the following example that shows that CENNET's architecture is effective for avoiding extracting too few causal explanatory variables. In preliminary experiments, when directly causal variables were inferred for the final output variable $Y$, we observed that only a few variables were selected as the CCVs for $Y$, making it difficult to provide sufficiently diverse causal explanatory power for individual prediction outcomes. For instance, we observed that the variable $X_{2}$, which was designed to be a direct causal variable for the prediction variable $Y$ itself, was not selected as a CCV for $Y$ in the Non-Linear Additive dataset, and similarly, the $X_{1}$ was also not selected as a CCV for $Y$ in the Category dataset. Therefore, we conclude the CCV selections for each neuron belonging to the NNLU are necessary at least in the selection of causal explanatory variables, and the method is consequently specialized for NNs. 

\section{Conclusion}
\label{conclusion}
In this study, we proposed Causal Explanations for Neural NETwork predictors (CENNET), a method to provide explanations from a causal perspective for NN predictors when we can assume that there are causal relationships behind a dataset in a tabular form. We introduced characteristic correlations and characteristic correlated variables (CCVs) and showed the excellent property of the CCVs for selecting variables from the viewpoint of Shannon entropy. Since this mechanism generates explanatory reasons from directly causal factors on the predicted results, it provides a way to move away from the possibility of users misunderstanding that there are causal reasons for given explanations generated from pseudo-correlated variables. At the same time, it is possible to eliminate indirect influencing factors, which may generate similar explanatory reasons to ones obtained from direct influencing factors and thus provide compact explanatory reasons.

CENNET enables us to globally understand the composition of complex features in a selected hidden layer of NNs, especially in the classification-type prediction task, made from the non-additive composition of input variables. CENNET also provides a causal explanation consisting of input factors that are combined for individual prediction results. Concerning this, CENNET can collectively evaluate reasons consisting of single and multiple factors with a proposed index. We evaluated and demonstrated the effectiveness of CENNET in some synthetic and quasi-real datasets. 

\section*{Acknowledgments}
We thank Ryosuke Nakajima and Junpei Tanikawa for their assistance with the preliminary literature review and data preparation in the study.

\bibliographystyle{unsrt}
\bibliography{cennet}

@article{GBM,
author = "J.~Friedman",
title = "Greedy Function Approximation: A Gradient Boosting Machine",
journal = "Annals of Statistics",
volume = "29",
issue = "5",
pages = "1189-1232",
year = 2001
}

@inproceedings{XGBoost,
author = "T.~Chen and C.~Guestrin",
title = "{XGB}oost: A Scalable Tree Boosting System",
booktitle = "Proceedings of the 22nd ACM SIGKDD International Conference on Knowledge Discovery 
and Data Mining",
pages = "785-794",
year = 2016,
}

@inproceedings{LightGBM,
author = "G.~Ke and Q.~Meng and T.~Finley and T.~Wang and W.~Chen and W.~Ma and Q.~Ye and T.Y.~Liu",
title = "Light{GBM}: A Highly Efficient Gradient Boosting Decision Tree",
booktitle = "Proceedings of the 31st Advances in Neural Information Processing Systems (NeurIPS 2017)",
pages = "3146-3154",
year = 2017
}

@inproceedings{LIME,
author = "M.~T.~Ribeiro and S.~Singh and C.~Guestrin",
title = "Why Should {I} Trust You? {E}xplaining the Predictions of Any Classifier",
booktitle = "Proceedings of the 22nd ACM SIGKDD International Conference on Knowledge Discovery 
and Data Mining",
pages = "1135-1144",
year = 2016
}

@inproceedings{SHAP,
author = "S.M.~Lundberg and S.I.~Lee",
title = "A Unified Approach to Interpreting Model Predictions",
booktitle = "Proceedings of the 31st Advances in Neural Information Processing Systems (NeurIPS 2017)",
pages = "4765-4774",
year = 2017
}

@inproceedings{NAM,
author = "R.~Agarwal and L.~Melnick and N.~Frosst and X.~Zhang and B.~Lengerich and R.~Caruana and G.~E.~Hinton",
title = "Neural Additive Models: Interpretable Machine Learning with Neural Nets",
booktitle = "Proceedings of the 35th Advances in Neural Information Processing Systems (NeurIPS 2021)",
year = 2021
}

@InProceedings{SLIME,
author =  "Z.~Zhou and G.~Hooker and F.~Wang",
title = 	 "{S-LIME}: Stabilized-LIME for Model Explanation",
booktitle = "Proceedings of the 27th ACM SIGKDD Conference on Knowledge Discovery and Data Mining",
pages =  "2429--2438",
year = 2021
}

@inproceedings{ACV,
 author = "S.~I.~Amoukou and N.~J-B.~Brunel",
  title = "Consistent Sufficient Explanations and Minimal Local Rules for Explaining the Decision of Any Classifier or Regressor",
 booktitle = "Proceedings of the 36th Advances in Neural Information Processing Systems (NeurIPS 2022)",
 year = 2022
}

@article{adadi2018peeking,
  title = "Peeking inside the black-box: a survey on explainable artificial intelligence ({XAI})",
  author = "A.~Adadi and M.~Berrada",
  journal = "IEEE Access",
  volume = "6",
  pages = "52138-52160",
  year = 2018
}

@article{IF_XAIconcepts,
title = "Explainable Artificial Intelligence ({XAI}): Concepts, taxonomies, opportunities and challenges toward responsible {AI}",
author = "A.B.~Arrieta and N.~D\'{i}az-Rodr\'{i}guez and J.D.~Ser and A.~Bennetot and S.~Tabik and A.~Barbado and S.~Garcia and 
S.~Gil-Lopez and D.~Molina and R.~Benjamins and R.~Chatila and F.~Herrera",
journal = "Information Fusion",
volume = "58",
pages = "82--115",
year = 2020
}

@inproceedings{kumar2020problems,
  title = "Problems with {S}hapley-value-based explanations as feature importance measures",
  author = "I.E.~Kumar and S.~Venkatasubramanian and C.~Scheidegger and S.~Friedler",
  booktitle = "Proceedings of the 37th International Conference on Machine Learning (ICML 2020)",
  year = 2020
}

@inproceedings{InterpretPredCausal-FLAIRS23,
title = "Interpreting Predictive Models through Causality: A Query-Driven Methodology",
author = "M.~H.~Ali and Y.~L.~Biannic and P-H.~Wuillemin",
booktitle = "Proceedings of the 36th International FLAIRS Conference",
year = 2023
}

@inproceedings{DecitionTree17DS,
title = "Deepred-rule Extraction from Deep Neural Networks",
author = "J.~R.~Zilke and E.~L.~Menc\'{i}a and F.~Janssen",
booktitle = "Proceedings of International Conference on Discovery Science",
pages = "457-473",
year = 2016,
}

@inproceedings{FrosstHintonDT17,
title = "Distilling a Neural Network into a Soft Decision Tree",
author = "N.~Frosst and G.~Hinton",
booktitle = "Proceedings of the First International Workshop on Comprehensibility and Explanation in AI and ML 2017 (CEX 2017)",
year = 2017,
}

@inproceedings{JanzingMinoricsBlobaumAISTATS20,
title = "Feature Relevance Quantification in Explainable {AI}: A Causal Problem",
author = "D.~Janzing and L.~Minorics and P.~Bl{\"o}baum",
booktitle = "Proceedings of the 23rd International Conference on Artificial Intelligence and Statistics (AISTATS 2020)",
pages = "2907-2916",
year = 2020,
}

@inproceedings{CXPlain19,
title = "{CXP}lain: Causal Explanations for Model Interpretation under Uncertainty",
author = "P.~Schwab and W.~Karlen",
booktitle = "Proceedings of the 33rd Conference on Neural Information Processing Systems (NeurIPS 2019)",
year = 2019,
}

@inproceedings{NNasSCM19,
title = "Neural Network Attributions: A Causal Perspective",
author = "A.~Chattopadhyay and P.~Manupriya and A.~Sarkar and V.~N.~Balasubramanian",
booktitle = "Proceedings of the 36th International Conference on Machine Learning (ICML 2019)",
pages = "981-990",
year = 2019,
}

@inproceedings{FryeFeigeRowat20nips,
title = "Asymmetric {S}hapley Values: Incorporating Causal Knowledge into Model-agnostic Explainability",
author = "C.~Frye and I.~Feige and C.~Rowat",
booktitle = "Proceedings of the 34th Conference on Advances in Neural Information Processing Systems (NeurIPS 2020)",
year = 2020,
}

@inproceedings{CausalShapley20nips,
title = "Causal Shapley Values: Exploiting Causal Knowledge to Explain Individual Predictions of Complex Models",
author = "T.~Heskes and E.~Sijben and I.~G.~Bucur and T.~Claassen",
booktitle = "Proceedings of the 34th Conference on Advances in Neural Information Processing Systems (NeurIPS 2020)",
year = 2020,
}

@article{IBM_DLCM,
title = "Explaining Deep Learning using Causal Inference",
author = "T.~Narendra and A.~Sankaran and D.~Vijaykeerthy and S.~Mani",
journal = "arXive preprint arXiv: 1811.04376",
year = 2018,
}

@inproceedings{WACV_causal_2024,
title = "Causal Analysis for Robust Interpretability of Neural Networks",
author = "O.~Ahmad and N.~Bereux and L.~Baret and V.~Hashemi and F.~Lecue",
booktitle = "Proceedings of IEEE/CVF Winter Conference on Applications of Computer Vision (WACV 2024)",
pages = "4673-4682",
year = 2024
}

@inproceedings{ICML2017best,
title = "Understanding Black-box Predictions via Influence Functions",
author = "P.~W.~Koh and P.~Liang",
booktitle = "Proceedings of the 34th International Conference on Machine Learning (ICML 2017)",
pages = "1885-1894",
year = 2017,
}

@inproceedings{TsangDetectInteraction,
title = "Detecting Statistical Interactions from Neural Network Weights",
author = "M.~Tsang and D.~Cheng and Y.~Liu",
booktitle = "Proceedings of International Conference on Learning Representations (ICLR 2018)",
year = 2018,
}

@inproceedings{selvaraju2017grad,
  title={Grad-cam: Visual explanations from deep networks via gradient-based localization},
  author={R.~R.~Selvaraju and M.~Cogswell and A.~Das and R.~Vedantam and D.~Parikh and D.~Batra},
  booktitle={Proceedings of IEEE international Conference on Computer Vision},
  pages={618--626},
  year={2017}
}

@inproceedings{ribeiro2018anchors,
  title={Anchors: High-precision model-agnostic explanations},
  author={M.~T.~Ribeiro and S.~Singh and C.~Guestrin},
  booktitle={Proceedings of the 23rd AAAI Conference on Artificial Intelligence},
  year={2018}
}

@InProceedings{pmlr-v80-chen18j,
  title = {Learning to Explain: An Information-Theoretic Perspective on Model Interpretation},
  author = {J.~Chen and L.~Song and M.~Wainwright and M.~Jordan},
  booktitle = {Proceedings of the 35th International Conference on Machine Learning (ICML2018)},
  pages = {883--892},
  year = 	 {2018}
}

@InProceedings{NeurIPS21DNNtabular,
title = "Revisiting Deep Learning Models for Tabular Data",
author = "Y.~Gorishniy and I.~Rubachev and V.~Khrulkov and A.~Babenko",
booktitle = "Proceedings of the 35th Conference on Advances in Neural Information Processing Systems (NeurIPS 2021)",
year = 2021,
}

@InProceedings{NeurIPS22WhyTreeBasedOutperform,
title = "Why Do Tree-Based Models still Outperform Deep Learning on Typical Tabular Data?",
author = "L.~Grinsztajn and E.~Oyallon and G.~Varoquaux",
booktitle = "Proceedings of the 36th Conference on Advances in Neural Information Processing Systems (NeurIPS 2022)",
year = 2022,
}

@article{NatureMI_Rudin,
title = "Stop Explaining Black Box Machine Learning Models for High Stakes Decisions 
and Use Interpretable Models Instead",
author = "C.~Rudin",
journal = "Nature Machine Intelligence",
volume = "1",
pages = "206--215",
year = 2019,
}

@article{DNN_survey_IEEE2022,
  title={Deep Neural Networks and Tabular Data: A Survey},
  author={V.~Borisov and T.~Leemann and K.~Se{\ss}ler and J.~Haug and M.~Pawelczyk and G.~Kasneci},
  journal={arXiv preprint arXiv:2110.01889},
  year={2021}
}

@InProceedings{ResNet,
author = "K.~He and X.~Zhang and S.~Ren and J.~Sun",
title = "Deep Residual Learning for Image Recognition",
booktitle = "2016 IEEE Conference on Computer Vision and Pattern Recognition",
pages = "770-778",
year = 2016,
}

@InProceedings{SNN,
author = "G.~Klambauer and T.~Unterthiner and A.~Mayr and S.~Hochreiter",
title = "Self-normalizing neural networks",
booktitle = "Proceedings of the 31th Conference on Advances in Neural Information Processing Systems (NeurIPS 2017)",
pages = "971-980",
year = 2017,
}

@InProceedings{KadraNeurIPS2021welltuned,
author = "A.~Kadra and M.~Lindauer and F.~Hutter and J.~Grabocka",
title = "Well-tuned Simple Nets Excel on Tabular Datasets",
booktitle = "Proceedings of the 35th Conference on Advances in Neural Information Processing Systems (NeurIPS 2021)",
year = 2021,
}

@inproceedings{TabNetAAAI21,
  title={Tab{N}et: {A}ttentive Interpretable Tabular Learning},
  author={S.~{\"O}.~Arlk and T.~Pfister},
  booktitle={Proceedings of the 35th AAAI Conference on Artificial Intelligence},
  year={2021}
}

@article{WhenNNoutperformGBDT23,
title = "When Do Neural Nets Outperform Boosted Trees on Tabular Data?",
author = "D.~McElfresh and S.~Khandagale and J.~Valverde and V.~Prasad~C. and 
B.~Feuer and C.~Hegde and G.~Ramakrishnan and M.~Goldblum and C.~White",
journal = "arXiv preprint arXiv: 2305.02997v3",
year = 2023
}

@article{tabular_NN24,
title = "Tabular Data: Is Attention All You Need?",
author = "G.~Zab{\"e}rgja and A.~Kadra and J.~Grabocka",
journal = "arXiv preprint arXiv: 2402.03970",
year = 2024
}

@inproceedings{Layer-Wise_NN,
title = "Layer-Wise Relevance Propagation for Neural Networks with Local Renormalization Layers",
author = "A.~Binder and G.~Montavon and S.~Bach and K.-R.~M{\"u}ller and W.~Samek",
booktitle = "Proceedings of the 25th International Conference on Artificial Neural Networks (ICANN 2016)",
year = 2016,
}

@article{PsyshologyCausal,
title = "Thinking Clearly About Correlations and Causation: Graphical Causal Models for Observational Data",
author = "J.~M.~Rohrer",
journal = "Advances in Methods and Practices in Psychological Science",
volume = "1",
issue = "1",
pages = "27--42",
year = 2018,
}

@article{MedicineCausal,
title = "Causality and Explainability of Artificial Intelligence in Medicine",
author = "A.~Holzinger and G.~Langs and H.~Denk and K.~Zatloukal and H.~M{\"u}ller",
journal = "WIREs Data Mining Knowledge Discovery",
volume = "9",
issue = "4",
doi = "https://doi.org/10.1002/widm.1312",
year = 2019,
}

@article{DNN_Causal,
title = "Causal Learning and Explanation of Deep Neural Networks via Autoencoded Activations",
author = "M.~Harradon and J.~Druce and B.~Ruttenberg",
journal = "arXiv preprint arXiv: 1802.00541v1",
year = 2018,
}

@book{CoverThomas,
author = "T.~M.~Cover and J.~A.~Thomas",
title = "Elements of Information Theory",
edition = "Second",
address = "Hoboken, NJ",
publisher = "John Wiley \& Sons",
year = "2006",
}

@book{LauritzenGM,
author = "S.~L.~Lauritzen",
title = "Graphical Models",
address = "New York, NY",
publisher = "Oxford University Press",
year = "1996",
}

@book{PearlCausality,
author = "J.~Pearl",
title = "Causality, models, reasoning, and inference",
address = "New York, NY",
publisher = "Cambridge University Press",
year = "2000",
}

@book{PC,
author = "P.~Spirtes and C.~Glymour and R.~Scheines",
title = "Causation, Prediction and Search",
edition = "Second",
address = "Cambridge, MA",
publisher = "MIT Press",
year = "2000",
}

@article{SpirtesIntro,
author = "P.~Spirtes",
title = "Introduction to Causal Inference",
journal = "Journal of Machine Learning Research",
volume = "11",
pages = "1643-1662",
year = 2010
}

@article{FGG,
author = "N.~Friedman and D.~Geiger and M.~Goldszmidt", 
title = "{B}ayesian Network Classifiers", 
journal = "Machine Learning", 
volume = "29", 
number = "2-3",
pages = "131-163", 
year = 1997,
}

@inproceedings{ALARM,
author = "I.~Beinlich and H.~Suermondt and R.~Chavez 
and G.~Cooper",
title = "The {ALARM} monitoring system: A Case Study 
with Two Probabilistic Inference Techniques for Belief Networks",
booktitle = "Proc. of European Conference on 
Artificial Intelligence in Medicine (AIME-89)",
pages = "247-256",
year = "1989",
}

@article{Hailfinder,
author = "B.~Abramson and J.~Brown and R.~L.~Winkler",
title = "Hailfinder: A {B}ayesian System for Forecasting 
Severe Weather",
journal = "International Journal of Forecasting",
volume = "12",
pages = "57-71",
year = "1996"
}

@article{Insurance,
author = "J.~Binder and D.~Koller and S.~Russell and K.~Kanazawa",
title = "Adaptive Probabilistic Networks with Hidden Variables",
journal = "Machine Learning",
volume = "29",
pages = "213-244",
year = "1997"
}

\section*{Appendix}
The appendix includes a background of probabilistic structural causal models, a proof of Theorem 1, and details of the experiments of CENNET compared with the existing XAI methods for the synthetic and quasi-real-world datasets. 
\setcounter{section}{0}
\section{Background of Probabilistic Structural Causal Models}\label{SCMbase}
A probabilistic structural causal model~\cite{PearlCausality,PC,SpirtesIntro} contains, in graphical representations, a node as a random variable and a directed edge as a direct influence between two nodes from a cause to an effect node. This graphical structure is a form of a directed acyclic graph (DAG). We describe several other graph-theoretical notions~\cite{PC,LauritzenGM}. A DAG $\mathcal{G}$ has a set of nodes $\bm{V}$ and a probability distribution $\mathsf{P}$. If there is a directed edge from node $A$ to node $B$, $A$ is a parent of $B$ and $B$ is a child of $A$, and we denote this by $A \in \text{PA}_{B}$ and $B \in \text{CH}_{A}$. A directed path from $A$ to $C$ (via $B$) is a sequence of directed edges such that $A \rightarrow B \rightarrow C$. If there is a directed path from $A$ to $C$, $C$ is a descendant of $A$, which is denoted by $C \in \text{DE}_{A}$, and $A$ is an ancestor of $C$. We also say that $A$ and $B$ precede $C$ in causal order when a causal path $A \rightarrow B \rightarrow C$ exists. The probability distribution $\mathsf{P}$ that entails a DAG then has the Markov condition~\cite{PC,LauritzenGM}: every node $X$ is independent of $\bm{V} \backslash (\text{DE}_{X} \cup \text{PA}_{X})$ given $\text{PA}_{X}$. Let $\text{ND}_{X}$ be a complementary set of $\text{DE}_{X}$, and then the Markov condition is also represented by $X \ci \,(\text{ND}_{X} \backslash \text{PA}_{X}) \,|\, \text{PA}_{X}$. When the edges denote causal relationships, we call the condition the causal Markov condition and use it as the axiom in the model inference~\cite{PC}. 

For a random variable $X$, let $\text{DI}\,(X)$ be a set of direct influential variables. Then, a structural equation model in a continuous variable set has the form (assumed linear model) of $Y=\sum_{X_{i}\in \text{DI}\,(Y)} \alpha_{i} X_{i} + \epsilon_{i}$, where $\alpha$ denotes coefficients and $\epsilon$ denotes noisy terms. For discrete random variable sets, a joint probability is decomposed into a product of conditional probabilities as $P( X_{1}, \dots, X_{n} ) = \prod_{i} P(X_{i} \,|\, \text{DI} \,(X_{i}))$, where $ \{ X_{1}, \dots, X_{n} \} $ denotes a set of random variables. In algorithms for finding SCMs, an edge that denotes direct influence is recognized when partial correlations are always not zero given any subset of variables. The statistical tests used are typically Fisher's z for linear Gaussian models with continuous variables and $\chi^{2}$ or $G^{2}$ tests with categorical variables~\cite{PC}. 
\section{Proof of Theorem 1}\label{ProofTheorem}
\setcounter{thm}{0}
\begin{thm}
In a given dataset $\mathcal{D}$ with a discrete random variable set $\bm{V}$, if there is a causal DAG for the set $\bm{V}$, the entropy of $X \in \bm{V}$ subject to the variable set CCVs of $X$ (CCV ($X$)), $H(X\,|\,\text{CCV}(X))$, achieves the lower limit of the conditional entropy given any set $\bm{S}$ such that $\text{CCV}(X) \subseteq \bm{S} \subseteq \bm{V}\backslash X$ and $X$ is not an ancestor of $\bm{S}$. That is,
\begin{equation}
\text{min} \, H(X\,|\,\bm{S}) = H(X\,|\,\text{CCV}(X)).
\label{CCV_entropy}
\end{equation}
\end{thm}
\begin{proof}[Proof]
Let $Y$ be a discrete random variable, and $\bm{Z}$ and $\bm{W}$ be a set of discrete random variables, respectively. According to information theory, $H(X) \geq H(X\,|\,Y)$, and the equality holds when $X$ and $Y$ are independent~\cite{CoverThomas}. Similarly, $H(X\,|\,\bm{Z}) \geq H(X\,|\,\bm{W}, \bm{Z})$ and the equality holds when $X$ and $\bm{W}$ are conditionally independent given $\bm{Z}$, which is due to the non-negativity of Kullback-Leibler divergence~\cite{CoverThomas}. Note that $\bm{S} = \text{ND}_{X}$ and $\text{CCV}(X)=\text{PA}_{X}$. Then, suppose $\bm{W} =\text{ND}_{X} \backslash \text{PA}_{X}$. If $\bm{W} = \emptyset$, then $H(X\,|\, \bm{S})=H(X \,|\,\text{CCV}(X))$. Otherwise, we can assume that $\bm{W}$ has $m$ elements where $m \geq 1$, and then $H(X\,|\,\text{CCV}(X), \bm{W}) = H(X\,|\,\text{CCV}(X))$ because of the Markov condition:\\
$X \ci \,(\text{ND}_{X} \backslash \text{PA}_{X})\,|\, \text{PA}_{X} = X \ci \bm{W}\,|\, \text{CCV} (X)$.
\end{proof}
\section{The Details of Experiments}\label{ExpApp}
In this section, we describe the details of experiments performed in this research on common experimental settings, and the synthetic and quasi-real-world datasets. 
\subsection{Common Experimental Settings}\label{setting}
We trained multi-layer perceptrons (MLPs) of neural networks (NNs) with two hidden layers each with ReLU activation using the Adam optimizer. 
We set the number of epochs as 100 and selected the model that achieved the best performance in the validation data.
To measure performance, we used areas under the precision-recall curve (PR-AUC).
For evaluating the performance of the trained NN models, we also trained XGboost~\cite{XGBoost} which is a kind of ensemble learning machine known as a high-performance predictor, and confirmed that our trained NN models achieved a competitive PR-AUC score compared to XGboost in preliminary experiments. 

The number of samples for the experiments was 10,000 and each dataset was split into parts of training, validation, and testing. 
Their proportions were 80:10:10 in the synthetic datasets and 90:5:5 in the Quasi-real datasets\footnote{Due to the high computational cost of ACV, the number of test data was set to 500 for the Quasi-real data experiment.}.
The computing environment used for running experiments was as follows. CPU: Intel Core i5, 2.30GHz, GPU: not used, main memory: 8GB, and OS: Windows 10. We implemented codes with Python 3 and used PyTorch 1.6.0 for NN models. 

In the comparative evaluation of CENNET and the existing XAI methods (LIME, S-LIME, SHAP, and ACV), 
we used the official Python implementations for LIME\footnote{https://github.com/marcotcr/lime}, S-LIME\footnote{https://github.com/ZhengzeZhou/slime}, SHAP\footnote{https://github.com/slundberg/shap} and ACV\footnote{https://github.com/salimamoukou/acv00} to produce the experimental results. 
For SHAP, we used Kernel SHAP, and other parameters were set to default. 
We set random seeds as 42 for NN model training and data splitting.
For causal inference, we used $G^{2}$ statistics with 1$\%$ significant level. 
\begin{figure}[t]
\centering
\includegraphics[bb = 0.000 0.000 1057.000 920.000, width = 9.0cm]{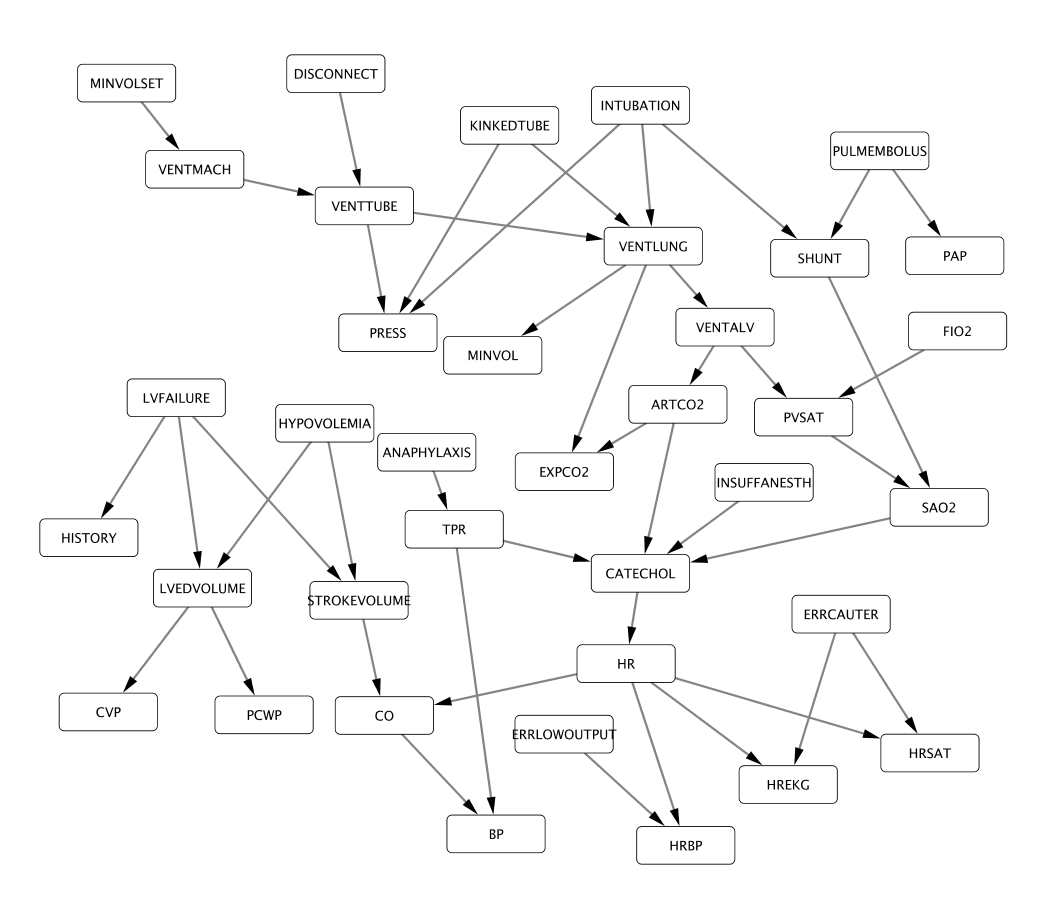}
\caption{The Alarm-DAG model for an example used in the Quasi-real data experiments.}
\label{Alarm}
\end{figure}
\subsection{Synthetic Datasets}
All the synthetic datasets used in the Non-Linear Additive, Non-Linear Non-Additive, and Category experiments were created by us because we could not find publicly available datasets with non-linear and non-additive relationships in that we wanted to validate XAI methods, to the best of our knowledge. 
\subsection{Details of Quasi-real-world Datasets}\label{QRWD}
Even in the experiment with real datasets, causal relationships should be known a priori for the evaluation of CENNET. 
We thus considered expert-created DAG models to be the closest to this condition and sampled the data from the models by considering the following DAG models as the SCMs. 
As such models, we used four expert knowledge DAG models: Alarm~\cite{ALARM}, Hailfinder~\cite{Hailfinder}, Insurance~\cite{Insurance}, and Carpo\footnote{https://www.cs.huji.ac.il/w$~$galel/Repository/Datasets/carpo/carpo.htm}. 
Figure~\ref{Alarm} shows the Alarm model with 37 variables as an example. 
We selected binary variables as the prediction variables for MLP predictors in each model. For those variables, parent variables are recognized in each DAG as shown in Table \ref{realdata_parent}. 
We selected candidate variables for the experiment, which were direct and indirect causal, pseudo-correlated, and resultant variables for the prediction variables. 
For instance, in the setting of a prediction variable CATECHOL in the Alarm model, we selected candidate variables as follows. As found in Figure~\ref{Alarm}, the set $\{$BP, EXPCO2$\}$ is the pseudo-correlated variable group, $\{$ARTCO2, SAO2, TPR$\}$ is the parent (i.e., direct causal) variable group (in which the variables were limited to three), $\{$ANAPHYLAXIS, PVSAT, VENTALV, SHUNT, PULMEMBOLUS, INTUBATION, VENTLUNG, FIO2$\}$ is the indirect causal variable group, and $\{$HR$\}$ is the resultant variable. We restricted the candidates to ancestors three generations back of the prediction variable in each setting. 
The experiments showed whether the XAI methods were able to select only direct causal variables as important ones from the candidates. The results are shown in Table \ref{res_realdata} of the paper. 

Table \ref{realdata_parent} shows the computational time in the Quasi-real data experiment for 500 test samples. 
Bold letters represent the best results examined by t-tests with 1$\%$ significance level in each setting. 
We made preliminary calculations of conditional probabilities as elements of EEP values and cached them for CENNET with training data only. 
The results show that the computation of CENNET is faster than the existing methods in nine out of ten settings.
\begin{table*}[hbt]
\vspace{0.3cm}
\caption{The parent variables (Parent Vars.) of each of the prediction variables in Quasi-real data experiments, and computational time (in seconds) in Quasi-real data experiments for 500 samples. Bold letters represent the best results in each setting.}
\begin{center}
\scalebox{0.95}[0.95]{
\small{
\begin{tabular}{l|l|ccccc}
\hline
\multicolumn{2}{c}{}  &  \multicolumn{5}{|c}{Computation Time (sec.)}\\ \hline
Dataset (Prediction Variable)&    $\#$/Names of Parent Vars. & LIME  & S-LIME & SHAP & ACV & CENNET  \\ \hline
Alarm (SHUNT) &  2 / PULMEMBOLUS, INTUBATION&24.41 & 134.09 & 139.96& 8137.86 & $\textbf{3.60}$ \\ 
Alarm (HISTORY) &1 / LVFAILURE& 22.63 & 52.33 & 17.87 & 247.80 & $\textbf{0.60}$ \\
Alarm (CATECHOL) &3 / ARTCO2, SAO2, TPR& $\textbf{27.67}$ & 81.91 & 269.83 &6001.50  & 35.37 \\
Hailinder (WindFieldMt) &1 / Scenario& 28.22 &150.32&267.68 & 5666.07 & $\textbf{1.38}$ \\ 
Hailfinder (ScenRelAMCIN) &1 / Scenario&28.55 & 128.81 & 243.78 & 4598.41 & $\textbf{0.69}$ \\ 
Insurance (Othercar) &1 / SocioEcon& 42.74 & 186.15 & 378.62 & 4618.55 & $\textbf{1.13}$ \\ 
Insurance (Vehicleyear) &2 / SocioEcon, RiskAversion&41.23   & 240.29 & 385.76 & 4974.11 & $\textbf{5.33}$ \\ 
Insurance (Antilock) &2 / VehicleYear, MakeModel & 42.82 & 186.00 & 218.06 & 5529.58 & $\textbf{5.72}$ \\ 
Insurance (Airbag) &2 / VehicleYear, MakeModel& 38.63 &198.72  & 58.43 & 1328.37 &  $\textbf{2.77}$ \\
Carpo (N42) &3 / N19, N27, N29 & 35.74 & 80.35 & 365.93 & 9543.55 & $\textbf{17.21}$ \\ \hline
\end{tabular}
}
}
\label{realdata_parent}
\end{center}
\end{table*}

\end{document}